\documentclass{llncs}
\usepackage{graphicx}
\usepackage{amsmath,amssymb} 
\usepackage{color}
\usepackage{multirow}
\usepackage[width=122mm,left=12mm,paperwidth=146mm,height=193mm,top=12mm,paperheight=217mm]{geometry}
\graphicspath{{./Figures/}}
\DeclareGraphicsExtensions{.pdf,.jpg,.png,.gif,.eps}

\DeclareMathOperator{\SLED}{SLED}

\DeclareMathOperator{\ARR}{ARR}
\DeclareMathOperator{\red}{red}
\DeclareMathOperator{\green}{green}
\DeclareMathOperator{\blue}{blue}
\begin{document}

\title{Efficient texture retrieval using multiscale local extrema descriptors and covariance embedding} 



\author{Minh-Tan Pham}
\institute{IRISA - Universit\'e Bretagne Sud, UMR 6074, F-56000, Vannes, France}

\maketitle

\begin{abstract}
We present an efficient method for texture retrieval using multiscale feature extraction and embedding based on the local extrema keypoints. The idea is to first represent each texture image by its local maximum and local minimum pixels. The image is then divided into regular overlapping blocks and each one is characterized by a feature vector constructed from the radiometric, geometric and structural information of its local extrema. All feature vectors are finally embedded into a covariance matrix which will be exploited for dissimilarity measurement within retrieval task. Thanks to the method's simplicity, multiscale scheme can be easily implemented to improve its scale-space representation capacity. We argue that our handcrafted features are easy to implement, fast to run but can provide very competitive performance compared to handcrafted and CNN-based learned descriptors from the literature. In particular, the proposed framework provides highly competitive retrieval rate for several texture databases including 94.95$\%$ for MIT Vistex, 79.87$\%$ for Stex, 76.15$\%$ for Outex TC-00013 and 89.74$\%$ for USPtex.
\end{abstract}

\section{Introduction}
Content-based image retrieval (CBIR) has been always drawing attention from researchers working on image analysis and pattern recognition within computer vision field. Texture, i.e. a powerful image feature involving repeated patterns which can be recognized by human vision, plays a significant role in most of CBIR systems. Constructing efficient texture descriptors to characterize the image becomes one of the key components which have been focused in most research works related to texture image retrieval \cite{tyagi2017content,raghuwanshi2016survey,alzu2015semantic}. 

From the literature, a great number of multiscale texture analysis methods using probabilistic approach have been developed to tackle retrieval task. In \cite{do2002wavelet}, the authors proposed to model the spatial dependence of pyramidal discrete wavelet transform (DWT) coefficients using the generalized Gaussian distributions (GGD) and the dissimilarity measure between images was derived based on the Kullback-Leibler divergences (KLD) between GGD models. Sharing the similar principle, multiscale coefficients yielded by the discrete cosine transform (DCT), the dual-tree complex wavelet transform (DT-CWT), the Gabor Wavelet (GW), etc. were modeled by different statistical models such as GGD, the multivariate Gaussian mixture models (MGMM), Gaussian copula (GC), Student-t copula (StC), or other distributions like Gamma, Rayleigh, Weibull, Laplace, etc. to perform texture-based image retrieval \cite{kwitt2008image,verdoolaege2008multiscale,choy2010statistical,kwitt2011efficient,lasmar2014gaussian,li2017color,yang2018weibull}. However, one of the main drawbacks of these techniques is the their expensive computational time which has been observed an discussed in several papers \cite{kwitt2011efficient,lasmar2014gaussian,li2017color}. 

Other systems which have provided effective CBIR performance include the local pattern-based framework and the block truncation coding (BTC)-based approach. The local binary patterns (LBP) were first embedded in a multiresolution and rotation invariant scheme for texture classification in \cite{ojala2002multiresolution}. Inspired from this work, many studies have been developed for texture retrieval such as the local maximum edge binary patterns (LMEBP) \cite{subrahmanyam2012local}, local ternary patterns (LTP) \cite{tan2010enhanced}, local tetra patterns (LTrP) \cite{murala2012local}, local tri-directional patterns (LTriDP) \cite{verma2016local}, local neighborhood difference pattern (LNDP) \cite{verma2018local}, etc. These descriptors, in particular LTrP and LTRiDP, can provide good retrieval rate. However, due to the fact that they work on grayscale images, their performance on natural textures is limited without using color information. To overcome this issue, recent schemes have proposed to incorporate these local patterns with color features. Some techniques can be mentioned here are the joint histogram of color and local extrema patterns (LEP+colorhist) \cite{murala2013joint}, the local oppugnant color texture pattern (LOCTP) \cite{jacob2014local}, the local extrema co-occurrence pattern (LECoP) \cite{verma2015local}, LBPC for color images \cite{singh2018color}. Beside that, many studies have also developed different BTC-based frameworks, e.g. the ordered-dither BTC (ODBTC) \cite{guo2013image,guo2015content}, the error diffusion BTC (EDBTC) \cite{guo2015content-based} and the dot-diffused BTC (DDBTC) \cite{guo2015effective}, which have provided competitive retrieval performance. Within these approaches, an image is divided into multiple non-overlapping blocks and each one is compressed into the so-called color quantizer and bitmap image. Then, a feature descriptor is constructed using the color histogram and color co-occurrence features combined with the bit pattern feature (including edge, shape, texture information) of the image. These features are extracted from the above color quantizer and bitmap image to tackle CBIR task.

Last but not least, not focusing on developing handcrafted descriptors as all above systems, learned descriptors extracted from convolution neural networks (CNNs) have been recently applied to image retrieval task \cite{schmidhuber2015deep,tzelepi2018deep}. An end-to-end CNN framework can learn and extract multilevel discriminative image features which are extremely effective for various computer vision tasks including recognition and retrieval \cite{schmidhuber2015deep}. In practice, instead of defining and training their own CNNs from scratch, people tend to exploit pre-trained CNNs (on a very large dataset such as ImageNet \cite{deng2009imagenet}) as feature extractors. Recent studies have shown the effective performance of CNN learned features w.r.t. traditional handcrafted descriptors applied to image classification and retrieval \cite{cusano2016evaluating,napoletano2017hand}.

In this work, we continue the traditional approach of handcrafted feature designing by introducing a powerful retrieval framework using multiscale local extrema descriptors and covariance embedding. Here, we inherit the idea of using local extrema pixels for texture description and retrieval from \cite{pham2017color} but provide a simpler and faster feature extraction algorithm which can be easily integrated into a multiscale scheme. Due to the fact that natural images usually involve a variety of local textures and structures which do not appear homogeneous within the entire image, an approach taking into account local features becomes relevant. Also, a multiscale approach could help to provide a better scale-space representation capacity to deal with complex textures. Within our approach, a set of local maximum and local minimum pixels (in terms of intensity) is first detected to represent each texture image. Then, to extract local descriptors, the image is divided into regular overlapping blocks of equal size and each block is characterized by a feature vector constructed using the radiometric (i.e. color), geometric ans structural information of its local extrema. The descriptor of each block is named SLED, i.e. simple local extrema descriptor. Thus, the input image is encoded by a set of SLED vectors which are then embedded into a feature covariance matrix. Moreover, thanks to the simplicity of the approach, we propose to upsample and downsample each image to perform the algorithm at different scales. Finally, we exploit the geometric-based riemannian distance \cite{FoMo2003} between covariance matrices for dissimilarity measurement within retrieval task. Our experiments show that the proposed framework can provide highly competitive performance for several popular texture databases compared against both state-of-the-art handcrafted and learned descriptors. In the rest of this paper, Section \ref{sec:method} describes the proposed retrieval framework including the details of SLED feature extraction, covariance embedding and multiscale scheme. We then present our experiments conducted on four popular texture databases in Section \ref{sec:experiments} and Section \ref{sec:conclusion} finally concludes the paper with some potential developments.

\section{Proposed texture retrieval framework}
\label{sec:method}
\subsection{Texture representation using local extrema pixels}
The idea of using the local extrema (i.e. local max and local min pixels) for texture analysis was introduced in \cite{PhMeMi2015,pham2014textural} for texture segmentation in very high resolution images and also exploited in \cite{pham2017color,pham2016texture2} for texture image retrieval. Regarding to this point of view, a texture is formed by a certain spatial distribution of pixels holding some illumination (i.e. intensity) variations. Hence, different textures are reflected by different types of pixel's spatial arrangements and radiometric variations. These meaningful properties can be approximately captured by the local extrema detected from the image. Hence, these local keypoints are relevant for texture representation and description \cite{pham2016change,pham2016pw,pham2016texture}. 
\begin{figure*}[ht]
\centering{
\begin{minipage}[b]{.48\linewidth}
  \centering
  \centerline{\includegraphics[width=55mm]{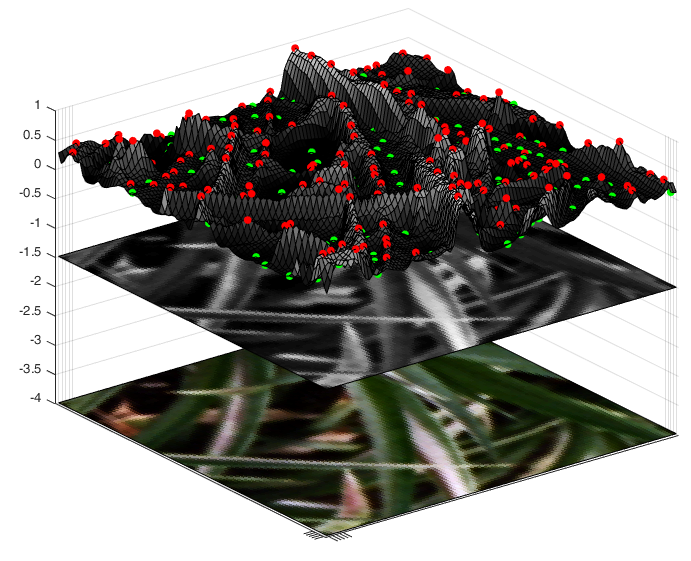}}              
  \footnotesize{(a) \emph{Leaves.0016}}
\end{minipage}
\begin{minipage}[b]{.48\linewidth}
  \centering
  \centerline{\includegraphics[width=55mm]{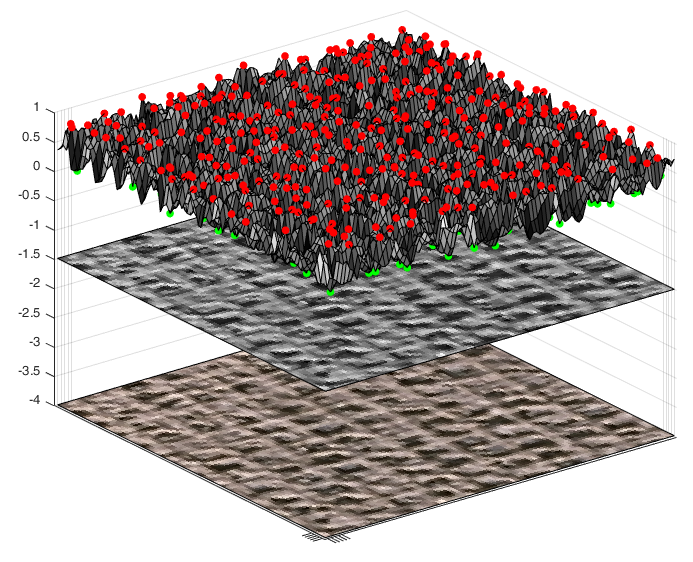}}              
  \footnotesize{(b) \emph{Fabric.0017}}
\end{minipage}
\vfill
\vspace{2mm}
\begin{minipage}[b]{.48\linewidth}
  \centering
  \centerline{\includegraphics[width=55mm]{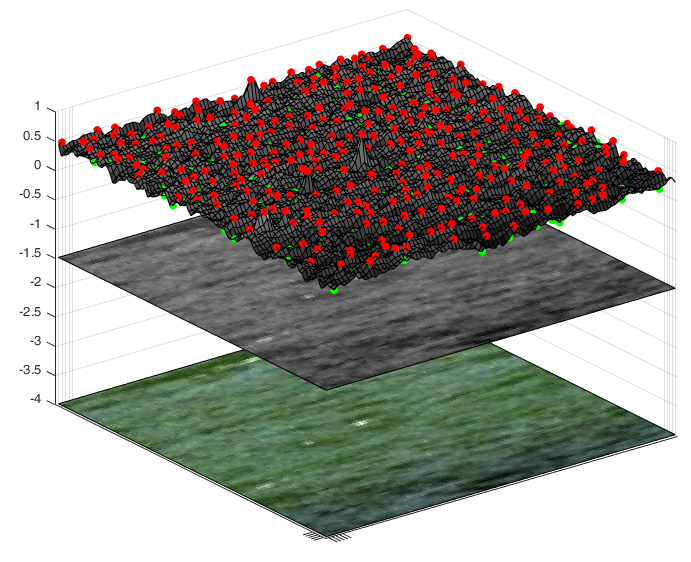}}              
  \footnotesize{(c) \emph{Water.0005}}
\end{minipage}
\begin{minipage}[b]{.48\linewidth}
  \centering
  \centerline{\includegraphics[width=55mm]{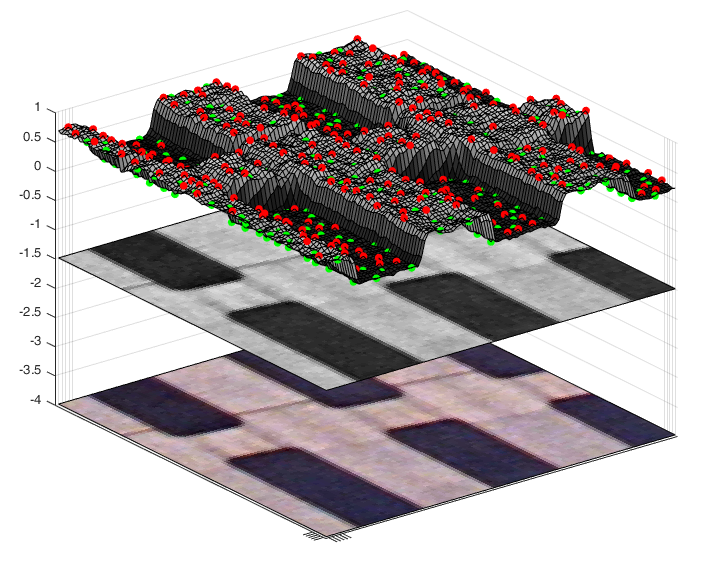}}              
  \footnotesize{(d) \emph{Buildings.0009}}
\end{minipage}
}
\caption{\textbf{Illustration:} spatial distribution and arrangement of local max pixels (red) and local min pixels (green) within 4 different textures from the MIT Vistex database \cite{vistex640}. These local extrema are detected using a $5\times 5$ search window.}
\label{fig:texture max-min}
\end{figure*}

The detection of local extrema from a grayscale image is quite simple and fast. Using a sliding window, the center pixel (at each sliding position) is supposed to be a local maximum (resp. local minimum) if it has the highest (resp. lowest) intensity value. Hence, by only fixing a $w \times w$ window size, the local extrema are detected by scanning the image only once. To deal with color images, there are different ways to detect local extrema (i.e. detecting from the grayscale version, using the union or intersection of extrema subsets detected from each color channel, etc.). For simplicity, we propose to detect local extrema from grayscale version of color images in this paper. To illustrate the capacity of the local extrema of representing and characterizing different texture contents, Fig. \ref{fig:texture max-min} shows their spatial apperance within 4 different textures of the MIT Vistex database \cite{vistex640}. Each $128\times 128$ color texture (at the bottom) is first converted to a grayscale image (in the middle). On the top, we display for each one a 3-D surface model using the grayscale image intensity as the surface height. The local maxima (in red) and local minima (in green) are detected using a $5\times 5$ sliding window. Some green points may be unseen since they are obscured by the surface. We observe that these extrema contain rich information that represent each texture content. Therefore, extracting and encoding their radiometric (or color) and geometric features could provide a promising texture description tool.
\subsection{Simple local extrema descriptor (SLED)}
\label{subsec:sled}
\begin{figure}
\centering
\centerline{\includegraphics[width=0.9\linewidth]{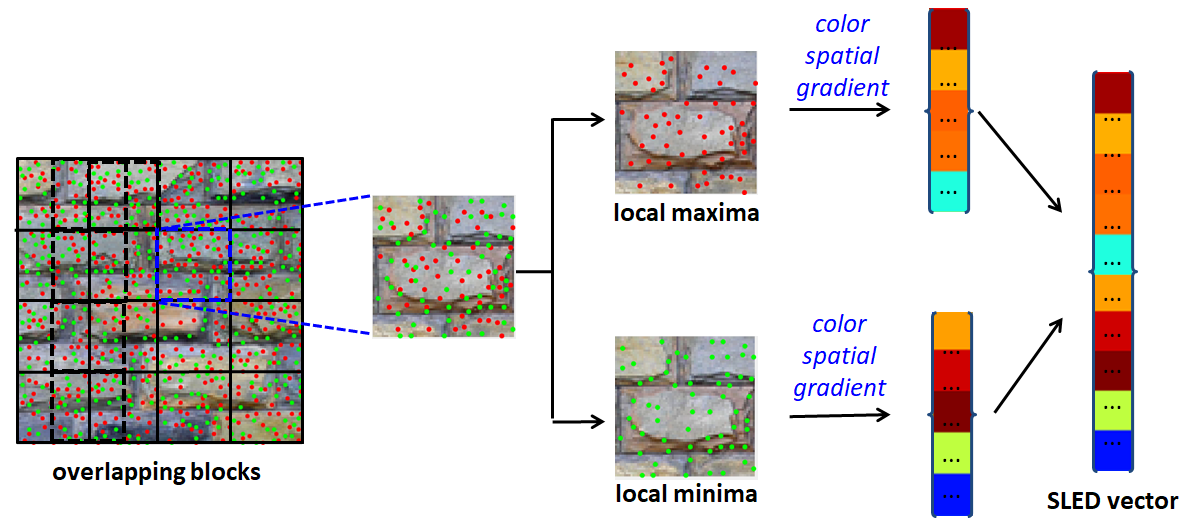}}
\caption{Generation of SLED feature vector for each image block.}
\label{fig:sled}
\end{figure}

Given an input texture image, after detecting the local extrema using a $w\times w$ sliding window, the next step is to divide the image into $N$ overlapping blocks of size $W\times W$ and then extract the simple local extrema descriptor (SLED) feature vector from each block. The generation of SLED vector is summarized in Fig. \ref{fig:sled}. From each image block $B_i, i=1\ldots N$, we first separate the local maxima set $S_i^{\max}$ and the local minima set $S_i^{\min}$, and then extract the color, spatial and gradient features of local keypoints to form their description vectors.

In details, below are the features extracted from $S_i^{\max}$, the feature generation for $S_i^{\min}$ is similar.
\begin{itemize}
\item[+] Mean and variance of each color channel:
 \begin{equation}
 \mu^{\max}_\text{i,color} =  \frac{1}{|S_i^{\max}|}\sum_{(x,y) \in S_i^{\max}}I_\text{color}(x,y),
 \label{eq:mean color}
 \end{equation}
 \begin{equation}
 \sigma_\text{i,color}^{2\max} = \frac{1}{|S_i^{\max}|}\sum_{(x,y) \in S_i^{\max}}(I_\text{color}(x,y)-\mu^{\max}_\text{i,color})^2,
 \label{eq:var color}
 \end{equation}
where $\text{color} \in \{ \red, \green, \blue \}$ represents each of the 3 color components; $(x,y)$ is the pixel position on the image grid and $|S_i^{\max}|$ is the cardinality of the set $S_i^{\max}$.

\item[+] Mean and variance of spatial distances from each local maximum to the center of $B_i$: 
 \begin{equation}
 \mu^{\max}_\text{i,spatial} = \frac{1}{|S_i^{\max}|}\sum_{(x,y) \in S_i^{\max}}d_i(x,y),
 \label{eq:mean distance}
 \end{equation}
 \begin{equation}
\sigma_\text{i,spatial}^{2\max} = \frac{1}{|S_i^{\max}|}\sum_{(x,y) \in S_i^{\max}}(d_i(x,y)-\mu^{\max}_\text{i,spatial})^2,
 \label{eq:var distance}
 \end{equation}
where $d_i(x,y)$ is the spatial distance from the pixel $(x,y)$ to the center of block $B_i$ on the image plane.
\item[+] Mean and variance of gradient magnitudes: 
 \begin{equation}
  \mu^{\max}_\text{i,grad} =\frac{1}{|S_i^{\max}|}\sum_{(x,y) \in S_i^{\max}}\nabla I(x,y),
 \label{eq:mean mag}
 \end{equation}
 \begin{equation}
 \sigma_\text{i,grad}^{2\max} = \frac{1}{|S_i^{\max}|}\sum_{(x,y) \in S_i^{\max}}(\nabla I(x,y)-\mu^{\max}_\text{i,grad})^2,
 \label{eq:var mag}
 \end{equation}
where $\nabla I$ is the gradient magnitude image obtained by applying the Sobel filter on the gray-scale version of the image.
\end{itemize}

All of these features are integrated into the feature vector $f_i^{\max} \in \mathbb{R}^{10}$, which encodes the color (i.e. three channels), spatial and structural features of the local maxima inside the block $B_i$:
\begin{equation} 
 	f_i^{\max} = \Big[
    \mu^{\max}_\text{i,color}, \sigma_\text{i,color}^{2\max}, \mu^{\max}_\text{i,spatial}, \sigma_\text{i,spatial}^{2\max}, \mu^{\max}_\text{i,grad}, \sigma_\text{i,grad}^{2\max}
 	\Big] \in \mathbb{R}^{10}.
\label{eq:led vector max}
\end{equation}

The generation of $f_i^{\min}$ from the local min set $S_i^{\min}$ is similar. Now, let $f_i^{\SLED}$ be the SLED feature vector generated for block $B_i$, we finally define:
\begin{equation} 
f_i^{\SLED} = \left[ f_i^{\max},f_i^{\min} \right] \in \mathbb{R}^{20}.
\label{eq:led vector}
\end{equation}

The proposed feature vector $f_i^{\SLED}$ enables us to characterize the local textures of each image block $B_i$ by understanding how local maxima and local minima are distributed and arranged, and how they capture color information as well as structural properties (given by gradient features). The extraction of our handcrafted SLED is quite simple and fast. We observe that it is also feasible to add other features to create more complex feature vector as proposed in \cite{pham2017color}. However, we argue that by using covariance embedding and performing multiscale framework (described in the next section), the simple and fast SLED already provides very competitive retrieval performance.

\subsection{Covariance embedding and multiscale framework}
\begin{figure*}
\centering
\centerline{\includegraphics[width=0.9\linewidth]{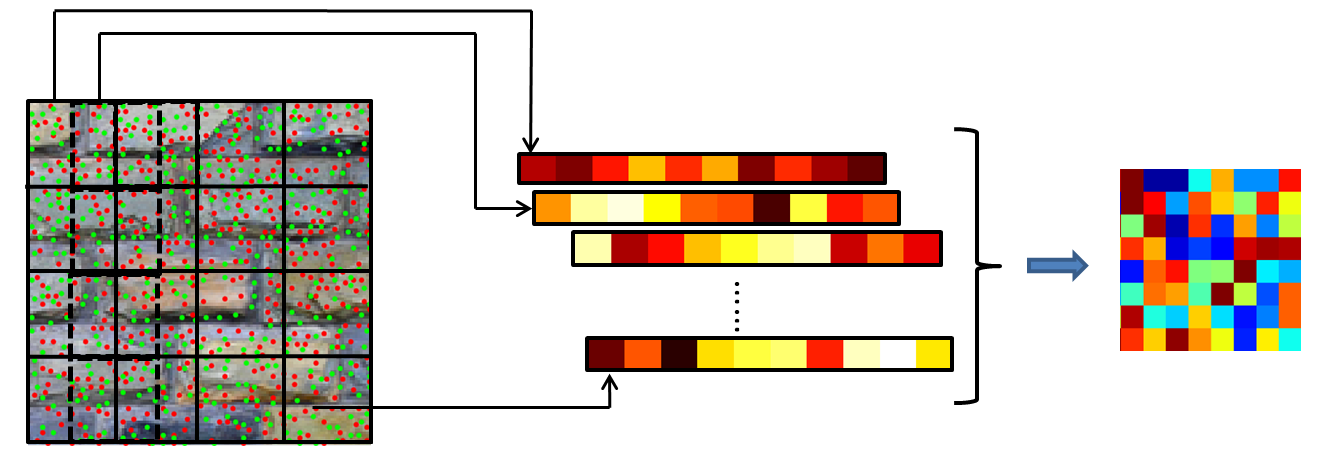}}
\caption{Proposed method to extract SLED feature vectors from all image overlapping blocks and embed them into a covariance matrix.}
\label{fig:embed}
\end{figure*}

The previous section has described the generation of SLED vector for each block of the image. Once all feature vectors are extracted to characterize all image blocks, they are embedded into a covariance matrix as shown in Fig. \ref{fig:embed}. Given a set of $N$ SLED feature vectors $f_i^{\SLED}, i=1\ldots N$, the embedded covariance matrix is estimated as follow:
\begin{equation}
C^{\SLED} = \frac{1}{N}\sum_{i=1}^N (f_i^{\SLED}-\mu^{\SLED})(f_i^{\SLED}-\mu^{\SLED})^T,
\label{eq:dcog}
\end{equation}
where $\mu^{\SLED} = \frac{1}{N}\sum_{i=1}^N f_i^{\SLED}$ is the estimated mean feature vector.

\begin{figure*}
\centering
\centerline{\includegraphics[width=0.9\linewidth]{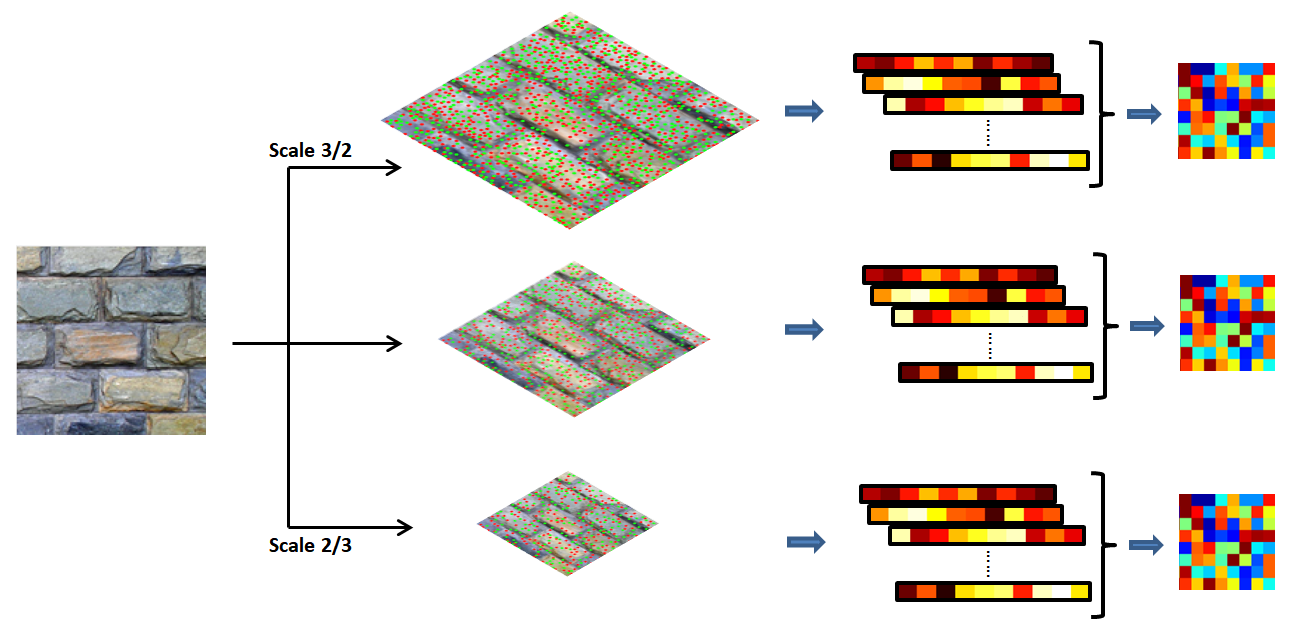}}
\caption{Proposed multi-scale framework.}
\label{fig:framework}
\end{figure*}

Last but not least, thanks to the simplicity of the proposed SLED extraction and embedding strategy, we also propose a multi-scale framework as in Fig. \ref{fig:framework}. Here, each input image will be upsampled and downsampled with the scale factor of 3/2 and 2/3, respectively, using the bicubic interpolation approach. Then, the proposed scheme in Fig. \ref{fig:embed} is applied to these two rescaled images and the original one to generate three covariance matrices. It should be noted that the number of rescaled images as well as scaling factors can be chosen differently. Here, without loss of genarality, we fix the number of scales to 3 and scale factors to 2/3, 1 and 3/2 for all implementations in the paper. To this end, due to the fact that covariance matrices possess a postitive semi-definite structure and do not lie on the Euclidean space, we finally exploit the geometric-based riemannian distance for dissimilarity measurement within retrieval task. This metric has been proved to be relevant and effective for covariance descriptors in the literature \cite{FoMo2003}.

\section{Experimental study}
\label{sec:experiments}

\subsection{Texture databases}
\label{subsec:data}
\begin{figure}[ht!]
\centering{
\begin{minipage}[b]{0.49\linewidth}
  \centering
  \centerline{\includegraphics[width = 45mm]{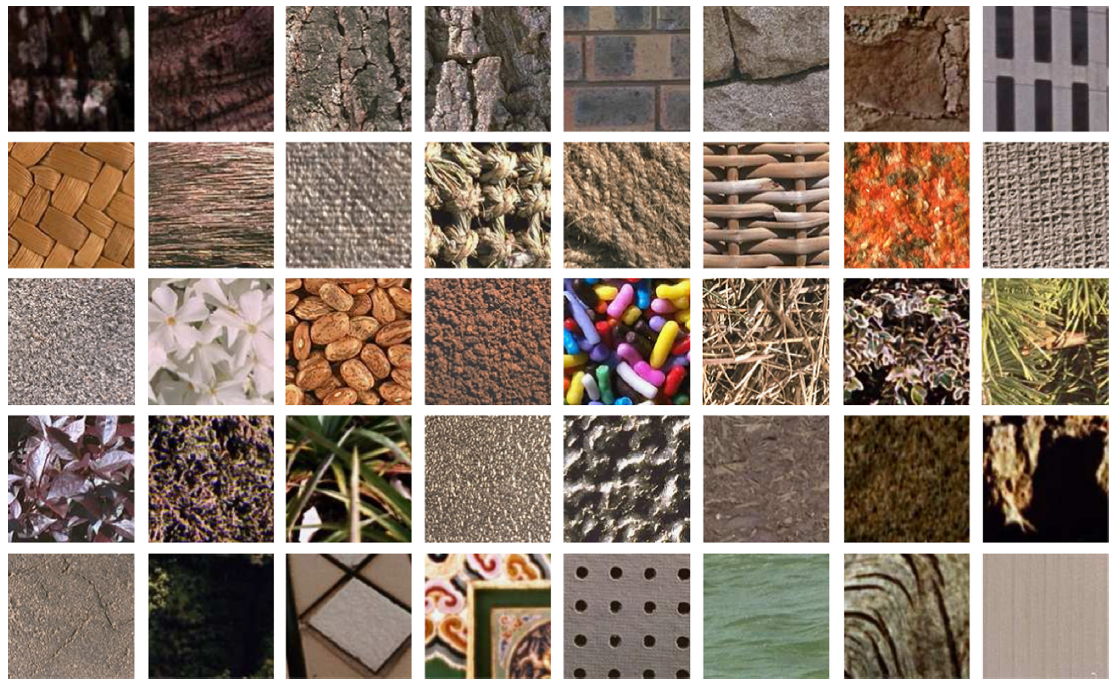}}              
  \footnotesize{(a) Vistex \cite{vistex640}}
\end{minipage}
\hfill
\begin{minipage}[b]{.49\linewidth}
  \centering
  \centerline{\includegraphics[width = 45mm]{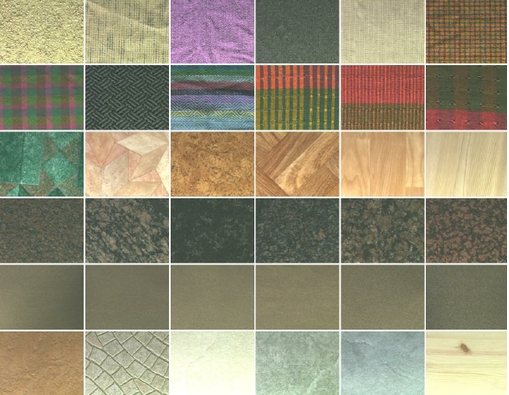}}              
  \footnotesize{(c) Outex \cite{outex-1360}}
\end{minipage}
\vfill
\vspace{2mm}
\begin{minipage}[b]{.49\linewidth}
  \centering
  \centerline{\includegraphics[width = 50mm]{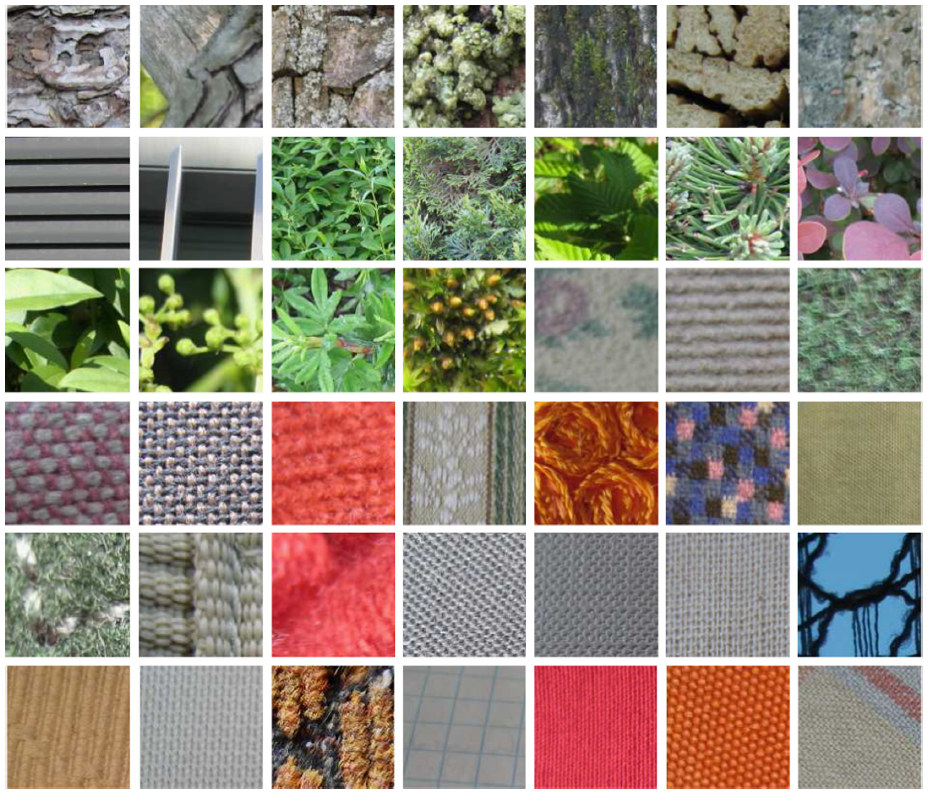}}              
  \footnotesize{(b) Stex \cite{stex7616}}
\end{minipage}
\hfill
\begin{minipage}[b]{.49\linewidth}
  \centering
  \centerline{\includegraphics[width = 50mm]{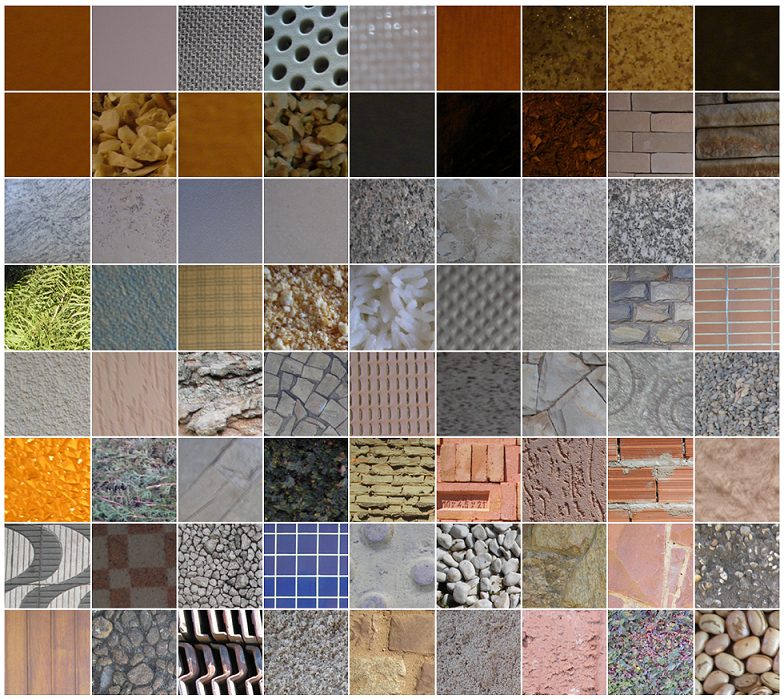}}              
  \footnotesize{(d) USPtex \cite{usptex-2292}}
\end{minipage}
}
\caption{Four image databases used in the experimental study.}
\label{fig:data}
\end{figure}

Four popular texture databases including the MIT Vistex \cite{vistex640}, the Salzburg Texture (Stex) \cite{stex7616}, the Outex TC-00013 \cite{outex-1360} and the USPtex \cite{usptex-2292} were exploited to conduct our experiments. Vistex is one of the most widely used texture databases for performance evalution and comparative study in the CBIR field. It consists of 640 texture images (i.e. 40 classes $\times$ 16 images per class) of size $128\times 128$ pixels. Being much larger, the Stex database is a collection of 476 texture classes captured in the area around Salzburg, Austria under real-word conditions. As for Vistex, each class includes 16 images of $128\times 128$ pixels, hence the total number of images from the database is 7616. The third dataset, the Outex TC-00013 \cite{outex-1360}, is a collection of heterogeneous materials such as paper, fabric, wool, stone, etc. It comprises 68 texture classes and each one includes 20 image samples of $128\times 128$ pixels. Finally, the USPtex database \cite{usptex-2292} includes 191 classes of both natural scene (road, vegetation, cloud, etc.) and materials (seeds, rice, tissues, etc.). Each class consists of 12 image samples of $128\times 128$ pixels. Fig. \ref{fig:data} shows some examples of each texture database and Table \ref{tab:database} provides a summary of their information.
\begin{table}[!ht]
	\centering
	\caption{Number of total images ($N_t$), number of classes ($N_c$) and number of relevant images ($N_R$) per class within the 4 experimental texture databases.}
	\label{tab:database}
	{\renewcommand{\arraystretch}{1.1}
	\resizebox{0.6\linewidth}{!}{
	\setlength\tabcolsep{6pt}
	\begin{tabular}{c | c c c c}	
	 & {\bf Vistex} & {\bf Stex} & {\bf Outex}  & {\bf USPtex}\\
	\hline
	$N_t$ & 640 & 7616 & 1380 & 2292 \\
	\hline
	$N_c$ & 40 & 476 & 68 & 191 \\
	\hline
	$N_R$ & 16 & 16 & 20 & 12\\
	\hline
	\end{tabular}
	}}
\end{table}

\subsection{Experimental setup}
\label{subsec:setup}
To perform our retrieval framework, the local extrema keypoints were detected using a $3\times 3$ search window ($w = 3$). We recommend this small window size to ensure a dense distribution of local extrema for different texture scenes. Next, each texture image is divided into overlapping blocks of size $32\times 32$ pixels and two consecutive blocks are $50\%$ overlapped. Thus, for each $128\times 128$ image, the number of blocks is $N=64$. For multiscale framework as in Fig.\ref{fig:framework}, we set 3 scales of 2/3, 1 and 3/2 as previously mentioned. There are no other parameters to be set, which confirms the simplicity of our proposed method.

For comparative evaluation, we compare our results to several state-of-the-art methods in the literature including:
\begin{itemize}
\item[+] probabilistic approaches in \cite{do2002wavelet,kwitt2008image,verdoolaege2008multiscale,choy2010statistical,kwitt2011efficient,lasmar2014gaussian},\cite{li2015rotation};
\item[+] handcrafted local pattern-based descriptors such as LMEBP \cite{subrahmanyam2012local}, LtrP \cite{murala2012local}, LEP+colorhist \cite{murala2013joint}, LECoP \cite{verma2015local}, ODII \cite{guo2013image};
\item[+] handcrafted BTC-based frameworks including DDBTC \cite{guo2015effective}, ODBTC \cite{guo2015content} and EDBTC \cite{guo2015content-based};
\item[+] learned descriptors based on pre-trained CNNs \cite{cusano2016evaluating,napoletano2017hand}. For these, we exploited the AlexNet \cite{krizhevsky2012imagenet}, VGG-16 and VGG-19 \cite{simonyan2014very} pre-trained on ImageNet database \cite{deng2009imagenet} as feature extractors. We used the 4096-D feature vector from the FC7 layer (also followed by a ReLu layer) and the L1 distance for dissimilarity measure as recommended in \cite{napoletano2017hand}.
\item[+] the LED framework proposed \cite{pham2017color} by setting equivalent parameters to our algorithm. In details, we set the 3 window sizes for keypoint extraction ($\omega_1$), local extrema detection ($\omega_2$) and LED generation ($W$) to $9\times 9$, $3\times 3$ and $36\times 36$, respectively.
\end{itemize}  
 
For evaluation criteria, the average retrieval rate (ARR) is adopted. Let $N_t$, $N_R$ be the total number of images in the database and the number of relevant images for each query, and for each query image $q$, let $n_q(K)$ be the number of correctly retrieved images among the $K$ retrieved ones (i.e. $K$ best matches). ARR in terms of number of retrieved images ($K$) is given by:
\begin{equation}
\ARR(K) = \frac{1}{N_t\times N_R}\sum_{q=1}^{N_t} n_q(K)\bigg|_{K \geq N_R}
\end{equation}

We note that $K$ is generally set to be greater than or equal to $N_R$. By setting $K$ equal to $N_R$, ARR becomes the primary benchmark considered by most studies to evaluate and compare the performance of different CBIR systems. All of ARR results shown in this paper were produced by setting $K = N_R$. 

\subsection{Results and discussion}
\label{subsec:result}
Tables \ref{tab:vistex_stex} and \ref{tab:usptex_outex} show the ARR performance of the proposed SLED and mutilscale SLED (MS-SLED) on our four texutre databases compared to reference methods. The first observation is that, most local feature-based CBIR schemes (e.g. LtrP \cite{murala2012local}, LEP+colorhist \cite{murala2013joint}, LECoP \cite{verma2015local}) or BTC-based techniques \cite{guo2015content,guo2015content-based,guo2015effective} have achieved better retrieval performance than probabilistic methods which model the entire image using different statistical distributions \cite{do2002wavelet,kwitt2008image,verdoolaege2008multiscale,choy2010statistical,kwitt2011efficient,lasmar2014gaussian}. Also, learned descriptors based on pre-trained CNNs have yielded very competitive retrieval rate compared to handcrafted features which prove the potential of CNN feature extractor applied to retrieval task \cite{cusano2016evaluating,napoletano2017hand}. Then, more importantly, our proposed SLED and MS-SLED frameworks have outperformed all reference methods for all datasets. We now discuss the results of each database to validate the effectiveness of the proposed strategy.
\begin{table}[!ht]
	\centering
	\caption{Average retrieval rate ($\%$) on the \textbf{Vistex} and \textbf{Stex} databases yielded by the proposed method compared to reference methods.}
	\setlength\tabcolsep{6pt}
	\begin{tabular}{l| r |r}	
	{\bf Method} & \textbf{Vistex} & \textbf{Stex}\\
	\hline
	GT+GGD+KLD \cite{do2002wavelet} & 76.57 & 49.30\\
	MGG+Gaussian+KLD \cite{verdoolaege2008multiscale}  & 87.40 & -\\
	MGG+Laplace+GD \cite{verdoolaege2008multiscale} & 91.70 & 71.30\\
	Gaussian Copula+Gamma+ML \cite{kwitt2011efficient} & 89.10 & 69.40\\
	Gaussian Copula+Weibull+ML \cite{kwitt2011efficient} & 89.50 & 70.60\\
	Student-t Copula+GG+ML \cite{kwitt2011efficient} & 88.90 & 65.60 \\
	Gaussian Copula+Gabor Wavelet \cite{li2017color} & 92.40 & 76.40\\
	LMEBP \cite{subrahmanyam2012local} & 87.77 & -\\
	LtrP \cite{murala2012local} & 90.02 & -\\
	LEP+colorhist \cite{murala2013joint} & 82.65 & 59.90\\
	DDBTC \cite{guo2015effective}  & 92.65 & 44.79\\
	ODBTC \cite{guo2015content}  & 90.67 & -\\
	EDBTC \cite{guo2015content-based}  & 92.55 & -\\
	LECoP \cite{verma2015local}  & 92.99 & 74.15\\
	ODII \cite{guo2013image}  & 93.23 & -\\
	CNN-AlexNet \cite{napoletano2017hand} & 91.34 & 68.84\\
	CNN-VGG16 \cite{napoletano2017hand}  & 92.97 & 74.92 \\
	CNN-VGG19 \cite{napoletano2017hand}  & 93.04 & 73.93\\
	LED \cite{pham2017color} & 94.13 & 76.71 \\	
	{\bf Proposed SLED}  & {\bf 94.31} & {\bf 77.78}\\
	{\bf Proposed MS-SLED}  & {\bf 94.95} & {\bf 79.87}\\
	\hline
	\end{tabular}
	\label{tab:vistex_stex}
\end{table}
\begin{table}[!ht]
	\centering	
	\caption{Average retrieval rate ($\%$) on the \textbf{Outex} and \textbf{USPtex} databases yielded by the proposed method compared to reference methods.}
	\label{tab:usptex_outex}
	\setlength\tabcolsep{6pt}
	\begin{tabular}{l |r| r}	
	{\bf Method} & {\bf Outex} & {\bf UPStex} \\
	\hline
	DDBTC ($L_1$) \cite{guo2015effective}  & 61.97 & 63.19\\
	DDBTC ($L_2$) \cite{guo2015effective}  & 57.51 & 55.38\\
	DDBTC ($\chi^2$) \cite{guo2015effective}  & 65.54 & 73.41\\
	DDBTC (Canberra) \cite{guo2015effective}  & 66.82 & 74.97\\
	CNN-AlexNet \cite{napoletano2017hand}  & 69.87 & 83.57\\
	CNN-VGG16 \cite{napoletano2017hand}  & 72.91 & 85.03\\
	CNN-VGG19 \cite{napoletano2017hand}  & 73.20 & 84.22\\	
	LED \cite{pham2017color} & 75.14 & 87.50 \\
	{\bf Proposed SLED}  & {\bf 75.96} & {\bf 88.60}\\
	{\bf Proposed MS-SLED}  & {\bf 76.15} & {\bf 89.74}\\
	\hline
	\end{tabular}
\end{table}

The best ARR of $94.95\%$ and $79.87\%$ was produced for Vistex and Stex by our MS-SLED algorithm. A gain of $0.82\%$ and $3.16\%$ was achieved compared to the second-best method with original LED features in \cite{pham2017color}. Within the proposed strategy, the multi-scale scheme has considerable improved the ARR from the single-scale SLED (i.e. $0.62\%$ for Vistex and $2.09\%$ for Stex), which confirms the efficiency of performing multiscale feature extraction and embedding for better texture description, as our motivation in this work. Next, another important remark is that most of the texture classes with strong structures and local features such as buildings, fabric categories, man-made object's surfaces, etc. were perfectly retrieved. Table \ref{tab:vistex perclass} shows the per-class retrieval rate for each class of the Vistex database. As observed, half of the classes (20/40 classes) were retrieved with $100\%$ accuracy (marked in bold). These classes generally consist of many local textures and structures. Similar behavior was also remarked for Stex data. This issue is encouraging since our motivation is to continue developing hand-designed descriptors which represent and characterize local features better than both handcrafted and learned descriptors from the literature.

Similarly, the proposed MS-SLED framework also provided the best ARR for both Outex ($76.15\%$) and USPtex data ($89.74\%$) (with a gain of $1.01\%$ and $2.24\%$, respectively), as observed in Table \ref{tab:usptex_outex}. Compared to learned descriptors based on pretrained AlexNet, VGG-16 and VGG-19, an improvement of $2.95\%$ and $4.71\%$ was adopted, which confirms the superior performance of our method over the CNN-based counterparts. To this end, the efficiency of the proposed framework is validated for all tested databases.

\begin{table}[!ht]
	\centering
	\caption{Per-class retrieval rate ($\%$) on the \textbf{Vistex-640} database using the proposed LED+RD method}
	\setlength\tabcolsep{6pt}
	\begin{tabular}{|l r|l r| l r|}	
	\hline
	\multicolumn{1}{|l}{\bf Class} & {\bf Rate} & {\bf Class} & {\bf Rate} & {\bf Class} & {\bf Rate} \\
	\hline
	Bark.0000 & 75.00 & Fabric.0015 & {\bf 100.00} &  Metal.0002 & {\bf 100.00} \\
	Bark.0006 & 94.14 & Fabric.0017 & 96.88 & Misc.0002 & {\bf 100.00} \\
	Bark.0008 & 84.38 & Fabric.0018 & {\bf 100.00} & Sand.0000 & {\bf 100.00} \\
	Bark.0009 & 77.73 & Flowers.0005 & {\bf 100.00} & Stone.0001 & 85.55 \\
	Brick.0001 & 99.61 & Food.0000 & {\bf 100.00} & Stone.0004 & 93.75\\
	Brick.0004 & 97.27 & Food.0005 & 99.61 & Terrain.0010 & 94.14 \\
	Brick.0005 & {\bf 100.00} & Food.0008 & {\bf 100.00} & Tile.0001 & 90.23 \\
	Buildings.0009 & {\bf 100.00} & Grass.0001 & 94.53 & Tile.0004 & {\bf 100.00} \\
	Fabric.0000 & {\bf 100.00} & Leaves.0008 & {\bf 100.00} & Tile.0007 & {\bf 100.00} \\
	Fabric.0004 & 78.13 & Leaves.0010 & {\bf 100.00} &  Water.0005 & {\bf 100.00} \\
	Fabric.0007 & 99.61 & Leaves.0011 & {\bf 100.00} & Wood.0001 & 98.44 \\
	Fabric.0009 & {\bf 100.00} & Leaves.0012 & 60.93 & Wood.0002 & 88.67 \\
	\cline{5-6}
	Fabric.0011 & {\bf 100.00} & Leaves.0016 & 90.23 & \multirow{2}{*}{\textbf{ARR}} & \multirow{2}{*}{\textbf{94.95}} \\
	Fabric.0014 & {\bf 100.00} & Metal.0000 & 99.21 &  & \\
	\hline
	\end{tabular}
	\label{tab:vistex perclass}
\end{table}
\begin{table}[!ht]
	\centering
	\caption{Comparison of feature vector length of different methods.}
	\label{tab:dimension}
	\setlength\tabcolsep{6pt}
	\begin{tabular}{l| c}	
	{\bf Method} & {\bf Feature dimension} \\
	\hline
	DT-CWT \cite{do2002wavelet}  & $(3\times 6+2)\times 2$=40 \\
	DT-CWT+DT-RCWT \cite{do2002wavelet}  & $2\times (3\times 6+2)\times 2$=80 \\
	LBP \cite{ojala2002multiresolution}  & 256 \\
	LTP \cite{tan2010enhanced} & $2\times 256 = 512$ \\
	LMEBP \cite{subrahmanyam2012local}  & $8\times 512 = 4096$ \\
	Gabor LMEBP \cite{subrahmanyam2012local} & $3\times 4\times 512 = 6144$ \\
	LEP+colorhist \cite{murala2013joint} & $16\times 8 \times 8 \times 8 = 8192$ \\
	LECoP($H_{18}S_{10}V_{256}$) \cite{verma2015local}  & $18+10+256 = 284$ \\
	LECoP($H_{36}S_{20}V_{256}$) \cite{verma2015local}  & $36+20+256 = 312$ \\
	LECoP($H_{72}S_{20}V_{256}$) \cite{verma2015local}  & $72+20+256 = 348$ \\
	ODII \cite{guo2013image} & 128+128 = 256 \\
	CNN-AlexNet \cite{napoletano2017hand}  & 4096 \\
	CNN-VGG16  \cite{napoletano2017hand} & 4096 \\
	CNN-VGG19  \cite{napoletano2017hand} & 4096 \\

	LED \cite{pham2017color} & $33\times(33+1)/2$ = 561 \\
	
	{\bf Proposed SLED}  &  $\pmb{20\times(20+1)/2 = 210}$\\
	{\bf Proposed MS-SLED}  &  $\pmb{3 \times 210 = 630}$\\
	\hline
	\end{tabular}	
\end{table}
Last but not least, Table \ref{tab:dimension} provides the comparison of descriptor dimensions within different methods. We note that our SLED involves a $20\times 20$ covariance matrix estimated as \eqref{eq:dcog}. Since the matrix is symmetrical, it is only necessary to store its upper or lower triangular entries. Thus, the SLED feature dimension is calculated as $20\times (20+1)=210$. For MS-SLED, we multiply this to the number of scales and hence the length becomes 630 in our implementation. Other feature lengths from the table are illustrated from their related papers. We observe that the proposed SLED has lower dimension than the standard LED in \cite{pham2017color} (i.e. 210 compared to 561) but can provide faster and better retrieval performance. To support this remark, we show in Table \ref{tab:time} a comparison of computational time for feature extraction and dissimilarity measurement of LED and SLED. In short, a total amount of 95.17 seconds is required by our SLED to run on the Vistex data, thus 0.148 second per image, which is very fast. All implementations were carried out using MATLAB 2017a on computer of 3.5GHz/16GB RAM.
\begin{table}[!ht]	
	\centering
	\caption{Computation time (in second) of LED and SLED feature extraction (FE) and dissimilarity measurement (DM). Experiments were conducted on the Vistex database.}
	\label{tab:time}
	\setlength\tabcolsep{6pt}
	\begin{tabular}{c | c c |c c |c c}	
	\multirow{2}{*}{\bf Version} & \multicolumn{2}{c}{\textbf{FE time}} & \multicolumn{2}{|c}{\textbf{DM time} (s)} & \multicolumn{2}{|c|}{\textbf{Total time}}\\
	\cline{2-7}
	 & $t_\text{data}$ & $t_\text{image}$ & $t_\text{data}$ & $t_\text{image}$ & $t_\text{data}$ & $t_\text{image}$ \\
	\hline
	LED \cite{pham2017color} & 193.77 & 0.308 & 21.39 & 0.033 & 215.16 & 0.336 \\
	SLED (ours) & 86.35 & 0.135 & 8.82 & 0.013 & 95.17 & 0.148\\
	\hline
	\end{tabular}
	\vfill
	\vspace{1mm}
	\footnotesize{$t_\text{data}$: time for the total database ; $t_\text{image}$: time per each image.}
\end{table}

\section{Conclusions}
\label{sec:conclusion}
We have proposed a simple and fast texture image retrieval framework using multiscale local extrema feature extraction and covariance embedding. Without chasing the current trends of deep learning era, we continue the classical way of designing novel handcrafted features in order to achieve highly competitive retrieval performance compared to state-of-the-art methodologies. The detection of local extrema as well as the extraction of their color, spatial and gradient features are quite simple but they are effective for texture description and encoding. We argue that the proposed MS-SLED does not require many parameters for tuning. It is easy to implement, fast to run and feasible to extend or improve. The best retrieval rates obtained for four texture benchmarks shown in our experimental study have confirmed the effectiveness of the proposed strategy. Future work can improve the performance of MS-SLED by exploiting other textural features within its construction. Also, we are now interested in integrating SLED features into an auto-encoder framework in order to automatically learn and encode richer information for better texture representation.

\bibliographystyle{splncs}
\bibliography{TanAbrv,egbib}
\end{document}